\documentclass[10pt,twocolumn,letterpaper]{article}

\usepackage{wacv}
\usepackage{times}
\usepackage{epsfig}
\usepackage{graphicx}
\usepackage{amsmath}
\usepackage{amssymb}

\usepackage{times}
\usepackage{epsfig}
\usepackage{graphicx}
\usepackage{amsmath}
\usepackage{amssymb}
\usepackage{multirow}
\usepackage{tabu}
\usepackage{subcaption}
\usepackage{xcolor}

\newcommand{\figref}[1]{Figure.~\ref{#1}}
\newcommand{\tabref}[1]{Table~\ref{#1}}
\newcommand{\eqnref}[1]{Eqn.~(\ref{#1})}
\newcommand{\secref}[1]{Sec.\ref{#1}}
\renewcommand{\vec}[1]{\mathbf{#1}}

\wacvfinalcopy 

\def\wacvPaperID{204} 

\ifwacvfinal\fi
\begin{document}

\title{Propose-and-Attend Single Shot Detector}

\author{Ho-Deok Jang \hspace{5mm} Sanghyun Woo \hspace{5mm} Philipp Benz \hspace{5mm} Jinsun Park \hspace{5mm} In So Kweon \\
Korea Advanced Institute of Science and Technology (KAIST)\\
{\tt\small \{hdjang, shwoo93, pbenz, zzangjinsun, iskweon\}@kaist.ac.kr}
}


\maketitle
\ifwacvfinal\thispagestyle{empty}\fi

\begin{abstract}
  We present a simple yet effective prediction module for a one-stage detector. The main process is conducted in a coarse-to-fine manner. First, the module roughly adjusts the default boxes to well capture the extent of target objects in an image. Second, given the adjusted boxes, the module aligns the receptive field of the convolution filters accordingly, not requiring any embedding layers. Both steps build a propose-and-attend mechanism, mimicking two-stage detectors in a highly efficient manner. To verify its effectiveness, we apply the proposed module to a basic one-stage detector SSD. Our final model achieves an accuracy comparable to that of state-of-the-art detectors while using a fraction of their model parameters and computational overheads. Moreover, we found that the proposed module has two strong applications. 1) The module can be successfully integrated into a lightweight backbone, further pushing the efficiency of the one-stage detector. 2) The module also allows \textit{train-from-scratch} without relying on any sophisticated base networks as previous methods do. 
\end{abstract}


\section{Introduction}
\label{sec:introduction}

Object detection has achieved significant advances with the introduction of convolutional neural networks (CNN). The recent detection frameworks can be divided into two categories: (\romannumeral 1) two-stage detectors~\cite{girshick2015fast,ren2015faster} and (\romannumeral 2) one-stage detectors~\cite{liu2016ssd,redmon2016yolo9000,lin2017focal}.

In two-stage detectors, the first stage proposes a sparse set of candidate object regions. After a feature pooling operation in the second stage, the proposed candidates are further classified and regressed. Two-stage detectors~\cite{girshick2015fast,ren2015faster,dai2016rfcn,he2017mask} have achieved top performance on several challenging benchmarks, such as PASCAL VOC~\cite{everingham2010pascal} or MS COCO~\cite{lin2014mscoco}. On the other hand, one-stage detectors directly classify and regress from the initial predefined default boxes. Recent one-stage detectors~\cite{liu2016ssd,redmon2016yolo9000} have achieved promising results with faster speed and lower memory-footprint. However, the accuracy of the one-stage detectors usually lags behind that of two-stage detectors~\cite{huang2017speed}.

We argue that this performance gap can be mainly attributed to an architectural limitation of the one-stage detectors,~\ie, the lack of the \textit{propose-and-attend} mechanism that is included in two-stage detectors. Due to the lack of this mechanism, one-stage detectors struggle with two main issues: 1) a heuristic box matching strategy, and 2) a mismatch between the receptive field of the prediction module and object-features.

During training, positive default boxes are selected only when their intersection over union (IoU) with their ground-truth box is above a certain threshold (\eg, 0.5). Thus, carefully setting the initial sizes and locations of the default boxes is crucial for the detection performance. Otherwise, an inferior initial default box configuration leads to few or imbalanced training samples. In two-stage approaches~\cite{ren2015faster,dai2016rfcn,he2017mask}, the issue is addressed by the region proposal step~\cite{ren2015faster} (\ie, \textit{propose mechanism}). However, one-stage detectors cannot handle this issue. Therefore, most approaches~\cite{liu2016ssd,redmon2016yolo9000,lin2017focal} use a large number of initial default boxes with varying scales and aspect ratios, which not only requires more parameters and computation overheads but also is heuristic. For this, we suggest a proposing process for one-stage detectors; it adjusts the initial default boxes to fit well with target objects in an adaptive manner. The process effectively imitates the region proposal step~\cite{ren2015faster} of two-stage detectors without stage-discrimination, ensuring high efficiency.

In addition to the adjustment of the default boxes, there is another issue of mismatch between the receptive field of the prediction module and the object-features proposed by the adjusted default boxes. The two-stage approaches handle the misalignment issue through a feature pooling operation (e.g., RoI pooling~\cite{girshick2015fast}), by which the prediction module can accurately attend to the object-features (\ie, \textit{attend mechanism}). However, one-stage detectors have no such operation for the 'attend' mechanism due to the \textit{fixed} receptive field of the prediction module regardless of the adjustment of the default boxes. To address this issue, we suggest an attending process for one-stage detectors; it modulates the receptive field of the prediction module according to the adjusted default boxes to accurately capture object-features.

By putting all together, we propose a novel prediction module, called propose-and-attend (PA) prediction module, for one-stage detectors. To demonstrate its effectiveness, we apply the module to a basic SSD framework~\cite{liu2016ssd} with a feature pyramid network backbone~\cite{lin2017fpn}, leading to our final detector called PASSD. The PASSD achieves a \textit{propose-and-attend} mechanism in the two-stage detector in an efficient manner. We empirically validate that our approach is a simple yet effective solution to significantly boost the detection performance of one-stage detectors with marginal parameters overhead. We also show that our prediction module can be successfully applied to a light-weight backbone and succeed in training from scratch.

\section{Related Work}

\noindent\textbf{Object Detection.}
Sliding-window approaches, in which a classifier is applied to a dense image grid, have dominated the pre-deep-learning era. However, since the arrival of deep learning, the conventional approaches have been replaced by convolutional neural network (CNN) based detectors. In particular, they can be divided into two main streams: two-stage and one-stage.

\noindent\textbf{Two-Stage Detectors.}
Two-stage detectors~\cite{ren2015faster,dai2016rfcn} are composed of two parts. The first part generates a sparse set of region proposals, and the second part further classifies and regresses the proposals. These two-stage detectors have occupied top entries of challenging benchmarks~\cite{dai2016rfcn,lin2017fpn,he2017mask}.

\noindent\textbf{One-Stage Detectors.}
OverFeat~\cite{sermanet2013overfeat} is one of the earliest one-stage detectors based on deep learning. Afterward, YOLO~\cite{redmon2016you} and SSD~\cite{liu2016ssd} were proposed with promising accuracy and real-time speed. RetinaNet~\cite{lin2017focal} further improves the accuracy by modifying the standard loss function, addressing the extreme class imbalance problem during training. However, we argue that they all suffer from the issues caused by the lack of the propose-and-attend mechanism, as mentioned in ~\secref{sec:introduction}. Recently, the RefineDet framework~\cite{zhang2018refinedet} suggests exploiting the default box refinement module to mimic the propose mechanism of the two-stage detectors. However, the RefineDet framework does not consider the way to provide diverse training samples during training phase, and it further misses the attend mechanism in its design; however, both are crucial for achieving high detection accuracy, as will be shown.

\noindent\textbf{Receptive Field.}
A previous study on the receptive field~\cite{luo2016understanding} shows that the size of the effective receptive field is much smaller than the theoretical one (i.e., resembling a 2D Gaussian shape). This implies that the mismatch between the receptive field of the prediction module and real object-features can lead to severe performance degradation. While two-stage detectors mitigate the problem via feature pooling operation~\cite{girshick2015fast}, one-stage detectors are prone to miss exact object-features. In this work, we resolve the issue with the efficient and effective attending process that enables the accurate extraction of object-features in the prediction module of one-stage detectors.


\begin{figure*}[t!]
    \centering
    \includegraphics[width=1\textwidth]{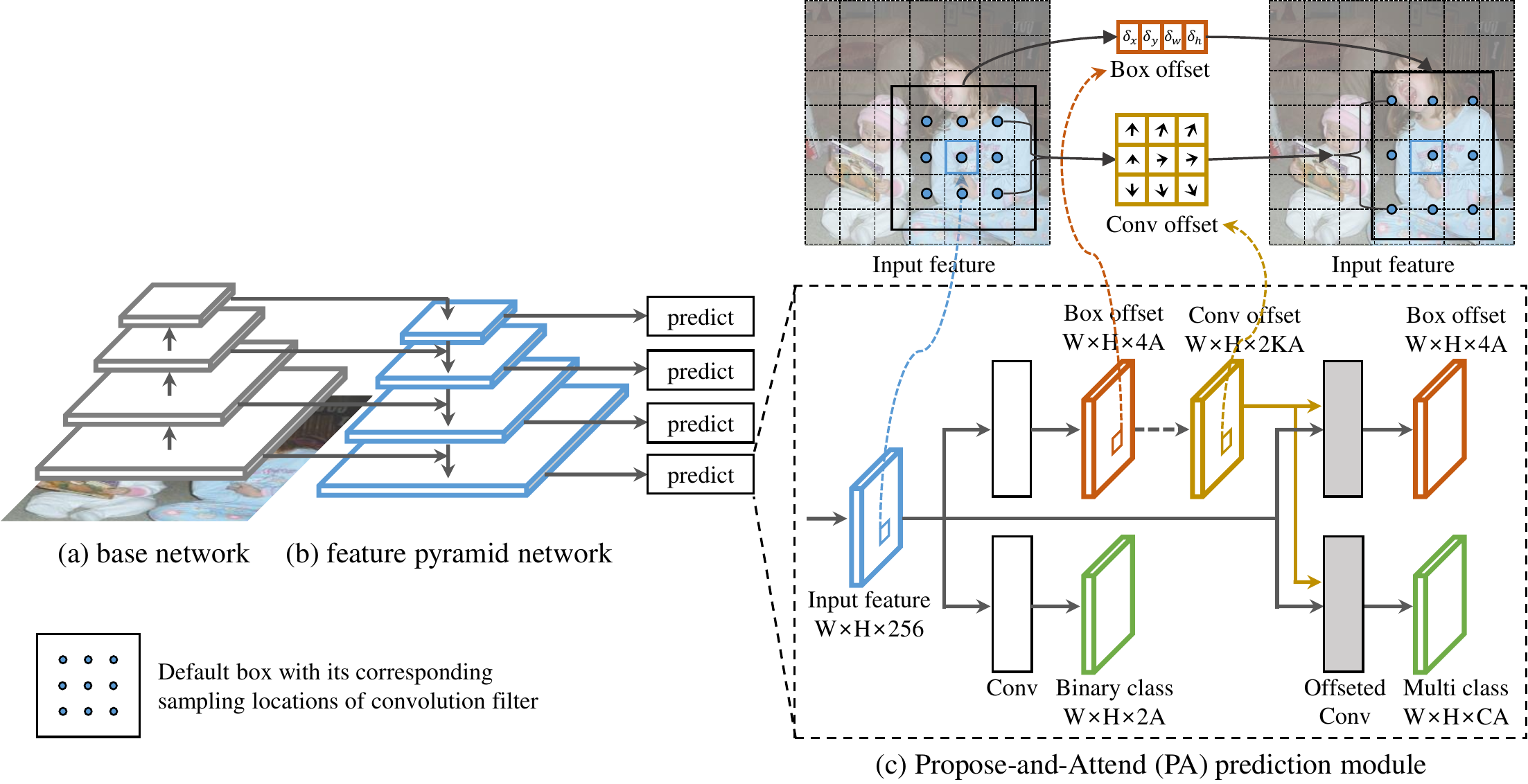}
    \caption{\textbf{Overall architecture of PASSD.} The model uses a feature pyramid network as a backbone network to generate a multi-scale feature pyramid. We then apply the propose-and-attend (PA) prediction module to each feature map to predict the final detection. The PA module adjusts the initial default boxes to capture the extent of the target objects and modulates the sampling locations of convolution filters accordingly to align the receptive field of the prediction module to the object-features. The whole pipeline is achieved in an end-to-end manner.}
    \label{fig:architecture}
\end{figure*}

\section{Method}
PASSD is a one-stage detection framework composed of an FPN~\cite{lin2017fpn} backbone network and the proposed propose-and-attend (PA) prediction module. The overall pipeline is shown in ~\figref{fig:architecture}. The backbone network generates multi-scale convolutional feature maps and is an off-the-shelf CNN. The PA prediction module produces final detection results based on the output of the backbone network. The final model features a simple yet effective design. We describe the details of the model in the following.


\section{Backbone Network}
To improve the scale-invariance of the model, we adopt the feature pyramid network (FPN)~\cite{lin2017fpn} as our backbone network. We apply the FPN on top of three base networks, VGG-16~\cite{simonyan2014very}, ResNet-101~\cite{he2016deep}, and MobileNet~\cite{howard2017mobilenets}. In order to capture large objects, we add two extra convolution blocks (\ie, conv8; stride=2, conv9; stride=2) to the end of the truncated VGG-16~\cite{simonyan2014very}, one extra bottleneck residual block (\ie, res6; stride=2, channel=512) to the end of the truncated ResNet-101~\cite{he2016deep}, and one extra depthwise convolution (\ie, stride=2, channel=512) to the end of the truncated MobileNet~\cite{howard2017mobilenets}, respectively. For the VGG-16 base network and its extra layers, we follow same configuration in SSD~\cite{liu2016ssd}. We also use the same L2 normalization technique to scale the feature norm of the VGG-16 following SSD~\cite{liu2016ssd}. To build the FPN, we use four~\footnote{For a model with an input size of 768, we use an additional feature map with a stride size of \{128\} pixels.} feature maps with the stride sizes of \{8, 16, 32, 64\} pixels from the base networks and their extra layers. Each feature map is selected right after the last layer that holds the corresponding stride size. We follow Lin et al.~\cite{lin2017fpn} for the details of the pyramid with a few minor modifications\footnote{We build a top-down pathway from the top extra layer added to the base networks, not from the top layer of base networks, for simplicity while maintaining accuracy.}. As in~\cite{lin2017fpn}, we use 256 channels for all pyramid levels.


\section{Default Boxes and Matching}
For each feature map from the backbone, we assign one default scale for the default boxes (\ie, 4 times the stride size of the corresponding feature map). For scale-variation of the objects, we associate each feature map cell with the default boxes with three scales \{$2^{0}$, $2^{1/3}$, $2^{2/3}$\}\footnote{For the MS COCO benchmark, we use \{$2^{-1/3}$, $2^{0}$, $2^{1/3}$\}.} of the default scale, and their aspect ratios are set to \{1:1\}. In total, we assign $A$=3 default boxes for each feature map cell. Each default box is responsible for detecting the object in it by predicting a length C vector for multi-class classification, where C is the number of object classes, including the background class, and four offsets for box regression, where offsets are encoded following the standard box parameterization~\cite{girshick2014rich}. During training, we assign the default box as positive if it has a Jaccard overlap score higher than 0.5 with the ground-truth box.

\smallskip
\subsection{Propose-and-Attend Mechanism}
\subsubsection{Proposing Process}
For the one-stage detectors, the initial configuration of the default boxes is crucial for the performance since they mostly rely on overlap-based training sample mining. However, the initial configuration is typically \textit{fixed}~\cite{liu2016ssd,fu2017dssd,lin2017focal}; thus, most one-stage detectors suffer from enumerating a considerable number of default boxes over the image space. This results in consuming exhaustive parameters and computations. On the other hand, two-stage detectors employ a region proposal step~\cite{ren2015faster}, which provides the model more abundant training samples by \textit{dynamically} offsetting the default boxes. However, its internal dependency on the proposal-wise computation significantly slows down the detection speed.

To overcome this issue, we design a proposing process that can function as the region proposal step~\cite{ren2015faster} of two-stage detectors without any proposal-wise computation (\eg, non-maximum suppression followed by top-k sorting). Specifically, the process adjusts the location and size of the default boxes to fit well with the target objects. To do this, we predict a binary objectness score that indicates the existence of a foreground object and four relative box offsets between the initial default box and the ground-truth following standard box parameterization~\cite{girshick2014rich} as follows:
\begin{align}\label{eqn:box_offset_parametrization}
  \begin{split}
      \delta_x = (g_x - b_x) / b_w, &
      \hspace{5mm}
      \delta_y = (g_y - b_y) / b_h \\
      \delta_w = \log{(g_w / b_w)}, &
      \hspace{5mm}
      \delta_h = \log{(g_h / b_h)}.
  \end{split}
\end{align}
Here, the default box is represented with its location (i.e., center) and size as $\vec{b} = (b_x, b_y, b_w, b_h)$, and the target ground-truth box as $\vec{g} = (g_x, g_y, g_w, g_h)$. The proposing process consists of one 3$\times$3 convolution layer with 2$A$ filters for binary objectness classification and the other one 3$\times$3 convolution layer with 4$A$ filters for box offsets regression ($A$=3 in this work). The parameters for the proposing process are shared across all the pyramid levels.

Moreover, we regularize the proposing process by clipping the offset regression as follows:
\begin{equation}\label{eqn:box_offsets_clipping}
  \delta_x' = {\tanh{(\delta_x)}} \cdot \frac{c_x}{2 b_w},\hspace{3mm} \delta_y' = {\tanh{(\delta_y)}} \cdot \frac{c_y}{2 b_h}
\end{equation}

Here, $c_x$ and $c_y$ are the stride sizes of each feature map cell. We observe that clipping operation induces discriminative feature learning during training. For example, without clipping, the positive training samples tend to focus on objects as a whole with high overlap between boxes, whereas clipping helps diversify the positive training samples to contain different parts of the objects. The result is similar with the application of non-maximum suppression in the region proposal step~\cite{ren2015faster} of two-stage detectors. We empirically confirm that this is important for improving the performance (refer to~\tabref{tab:PA_ablation}).

\subsubsection{Attending Process}
After the proposing process, the prediction module also necessitates attending on the adjusted default boxes (see \figref{fig:architecture}). In other words, the receptive field of the prediction module should be appropriately coordinated with the modified boxes to pick up object-features accurately. In two-stage detectors, a feature pooling operation (\eg, RoI pooling) is adopted to deal with this problem. However, the standard convolutional prediction module used in most one-stage detectors~\cite{liu2016ssd,lin2017focal,zhang2018refinedet} lacks this crucial attending operation in their design. Thus, valuable information for accurate classification and bounding box regression is lost.

To resolve the problem, we propose an efficient attending process. The main idea is to transform the sampling points of the convolution filter according to the modified default box. Before describing the details of the method, we first review the operation of the standard convolutional prediction module. For simplicity, we only describe the operation in 2D space without consideration of the channel axis, while extension to 3D is straightforward.

The classification or box regression for the $i$-th ($i\in\{1,\dots,A\}$) default box at the feature map location of $\vec{p}_o$ is computed by the weighted sum of the convolution filter ($\vec{w}$) and the input features ($\vec{x}$) on the sampling locations defined by sampling grid ($\mathcal{R}$) and $\vec{p}_o$, where $|\mathcal{R}|=k^{2}(K)$ and $k$ denotes the filter size of $k\times k$ convolution. An example case with a 3$\times$3 convolution filter and dilation of $1$ is shown below, where $\vec{p}_s$ enumerates the elements in $\mathcal{R}$:
$$\mathcal{R} = \{(-1,-1),(-1,0),\dots,(1,0),(1,1)\}$$
\begin{equation}
    \vec{y}^{i}(\vec{p}_o) = \sum_{\vec{p}_s \in \mathcal{R}} \vec{w}(\vec{p}_s) \cdot \vec{x}(\vec{p}_o + \vec{p}_s)
    \label{eqn:convolution}
\end{equation}

We can clearly observe that the sampling locations of the convolution filter are \textit{fixed} over the entire input feature map by the sampling grid ($\mathcal{R}$). This works fine under \textit{fixed} initial default boxes. However, with \textit{dynamically} adjusted default boxes, it becomes problematic. It misses the accurate object-features proposed by the adjusted default boxes, which are valuable for accurate classification and box regression. Moreover, it cannot explicitly take account of the adjusted default box in the final box regression. In other words, the box regressor is not aware of the target box it has to regress.

To address this problem, we instead augment the fixed grid sampling locations of the convolution filter with offsets that cover the adjusted default box in an adaptive manner. Specifically, the offsets ($\mathcal{O}_{o}^{i}$) for the $i$-th ($i\in\{1,\dots,A\}$) adjusted default box at the feature map location of $\vec{p}_o$ are obtained as:
$$\mathcal{O}_{o}^{i} = [ \mathcal{R} \odot (\hat{b}_{h}^{i}/k, \hat{b}_{w}^{i}/k) ] \oplus ({\Delta}y^{i}, {\Delta}x^{i})  - \mathcal{R} $$
\begin{equation}
    {\Delta}y^{i} = b_{h}^{i} \cdot \delta_{y}^{i}, \hspace{3mm} {\Delta}x^{i} = b_{w}^{i} \cdot \delta _{x}^{i}
    \label{eqn:center_displacement}
\end{equation}

Here, $(\hat{b}_{h}^{i}, \hat{b}_{w}^{i})$ and $({\Delta}y^{i}, {\Delta}x^{i})$ denote the size and center displacement of the adjusted default box, which are obtained by decoding the box parameterization (\eqnref{eqn:box_offset_parametrization}) and from~\eqnref{eqn:center_displacement}, respectively. Here, $\odot$ and $\oplus$ are element-wise multiplication and summation, respectively.

The given offsets ($\mathcal{O}_{o}^{i}$) allow the convolution filter to accurately capture the object-features in the adjusted default box (refer to~\figref{fig:architecture} for visual description where the offsets are denoted as 'Conv offset'):
\begin{equation}
  \vec{y}^{i}(\vec{p}_{o}) = \sum_{\vec{p}_{s} \in \mathcal{R}, \Delta\vec{p}_{s} \in \mathcal{O}_{o}^{i}} \vec{w}(\vec{p}_{s}) \cdot \vec{x}(\vec{p}_{o} + \vec{p}_s + {\Delta}\vec{p}_{s})
  \label{eqn:attending_process}
\end{equation}

Since standard convolution does not provide fractional sampling points, we implement offseted convolution following Dai et al.~\cite{dai2017deformable}. Unlike~\cite{dai2017deformable}, we enable multiple offsets for the output prediction to accommodate multiple default boxes at each feature map cell.

The attending process consists of one 3$\times$3 offseted convolution layer with $CA$ filters for multi-class classification including background (\ie, $C$ is 21 or 81 for PASCAL VOC~\cite{everingham2010pascal} and MS COCO~\cite{lin2014mscoco}) and the the other one 3$\times$3 offseted convolution layer with 4$A$ filters for the final box regression. The final box regression predicts the relative offsets between the adjusted default boxes and ground-truth boxes.

By putting all together, we build a novel prediction module for one-stage detectors, called propose-and-attend (PA) prediction module. The whole pipeline of the module is presented in~\figref{fig:architecture}. The PA prediction module significantly boosts one-stage detectors in efficient manner with marginal parameter overheads to the backbone network.

\section{Training and Inference}
\subsection{Training}

\medskip
\noindent\textbf{Data Augmentation.}
We follow several data augmentation strategies from SSD~\cite{liu2016ssd}. In brief, we use random photometric distortion, image flipping, and both zoom-in and zoom-out operations.

\medskip
\noindent\textbf{Hard Negative Mining.}
After the default box matching step, most of the default boxes are determined as negatives. To mitigate class imbalance, we use hard-negative mining following SSD~\cite{liu2016ssd}. We select hard-negative samples based on the loss values and constrain the ratio between positive and negative default boxes to be at most 1:3. We adopt this strategy for both the proposing and attending processes.

\medskip
\noindent\textbf{Loss Function.}
We define the total loss function as a sum of the two losses from the proposing and attending processes, respectively. Each loss term is formulated as follows:


$\mathcal{L} = \frac{1}{N_{pos}} ( \sum_{i}^{} \mathcal{L}_{cls} (p_i, c_i^*) + \sum_{i}^{} [x_i^* \geq 1] \mathcal{L}_{reg} (t_i, g_i^*) )$

\noindent Here, $i$ is the index of the default box in a mini-batch; $c_i^*$ is the ground truth class label of the default box $i$; $g_i^*$ are the ground truth default box offsets; $p_i$ and $t_i$ are the predicted class probability and box offsets of the default box $i$ in each process. $N_{pos}$ is the number of positive samples within the process in a mini-batch. If $N_{pos}$=0, we set the loss of the process to 0. The classification loss $\mathcal{L}_{cls}$ is cross-entropy loss, and we use Smooth-L1 loss~\cite{girshick2015fast} as the regression loss $\mathcal{L}_{reg}$. The Iverson bracket $[x_i^* \geq 1]$ outputs 1 when its condition is fulfilled, \ie, only the positive samples are included in the regression loss.

\medskip
\noindent\textbf{Optimization.}
For the initialization, we take the base networks (\ie, VGG-16, ResNet-101 and MobileNet) pre-trained on ImageNet. All new convolution layers are initialized using Gaussian weight with $\sigma = 0.01$ and bias $b = 0$. We set the batch size to 32~\footnote{For a model with an input size of 768, we use a 24 batch size due to the limited GPU.} during training. The entire network is trained using stochastic gradient descent (SGD) with a momentum of 0.9 and a weight decay of 0.0005. To stabilize the training process, we use a warmup strategy that gradually increases the learning rate from $10^{-6}$ to the initial learning rate of each dataset during the first 5 epochs. We use an initial learning rate of 4$\times10^{-3}$ for PASCAL VOC~\cite{everingham2010pascal} and 2$\times10^{-3}$ for MS COCO~\cite{lin2014mscoco}.

\medskip
\subsection{Inference}
PASSD predicts the final detection results in a fully convolutional manner. To ensure efficient inference, boxes with a score threshold lower than 0.01 are discarded and only the 200 top-scoring predictions per image are selected. Then, non-maximum suppression (nms) is applied to the top-scoring predictions with a threshold of 0.45 for duplication removal. For MS COCO, we use soft-nms~\cite{bodla2017soft} to filter out the boxes.

\begin{table*}[!t]
\centering
    \resizebox{0.93\textwidth}{!}{
    \begin{tabular}{c|c|c|c|c|c|c|ccc}
    \hline
    \multirow{4}{*}{exp} & \multicolumn{4}{c|}{prediction module} & \multirow{4}{*}{params} & \multirow{4}{*}{m$AP$} & \multirow{4}{*}{m$AP_{S}$} & \multirow{4}{*}{m$AP_{M}$} & \multirow{4}{*}{m$AP_{L}$}\\
    \cline{2-5} & \multicolumn{2}{c|}{\multirow{2}{*}{proposing process}} & \multicolumn{2}{c|}{attending process} & & & & & \\
    &\multicolumn{2}{c|}{*}& \multicolumn{2}{c|}{(convolution sampling locations)} &&&&&\\
    \cline{2-5}& w/o box offset clip & w/ box offset clip & fixed & adaptive & & & & & \\
    \hline
    \hline
    1 &               &              & Conv(dilation=1) &                 & 24.82M  & 75.5  & 48.8  & 72.6  & 85.5    \\
    2 & $\checkmark$  &              & Conv(dilation=1) &                 & 24.87M  & 79.0  & 51.8  & 77.1  & 86.2    \\
    3 &               & $\checkmark$ & Conv(dilation=1) &                 & 24.87M  & 79.8  & 54.6  & 78.2  & 86.0    \\
    \hline
    4 &               & $\checkmark$ & Conv(dilation=2) &                 & 24.87M  & 79.7  & 53.4  & 78.0  & 86.7    \\
    5 &               & $\checkmark$ & Conv(dilation=3) &                 & 24.87M  & 79.9  & 53.7  & 78.0  & 87.4    \\
    6 &               & $\checkmark$ &                  & Deform Conv     & 24.91M  & 80.2  & 55.0  & 78.4  & 86.9    \\
    \hline
    7(Ours)&          & $\checkmark$ &                  & Ours            & \textbf{24.87M}& \textbf{81.0} & \textbf{56.5} & \textbf{79.4} & \textbf{87.5} \\
    \hline
    \end{tabular}
    }
    \caption{Ablation experiments for propose-and-attend (PA) prediction module. We train all models on the union of VOC2007 \texttt{trainval} set and VOC2012 \texttt{trainval} set, and evaluate them on the VOC 2007 \texttt{test} set.}
\label{tab:PA_ablation}
\end{table*}

\begin{table}[t]
\centering
    \resizebox{\linewidth}{!}{
    \begin{tabular}{c|c|c|cccc}
    \hline
    \multicolumn{1}{c|}{attending process} & \multirow{1}{*}{params} & \multirow{1}{*}{m$AP_{.5}$} & \multirow{1}{*}{m$AP_{.6}$} & \multirow{1}{*}{m$AP_{.7}$} & \multirow{1}{*}{m$AP_{.8}$} \\
    \hline
    \hline
    Conv(dilation=1)   & 24.87M  & 79.8  & 74.3 & 63.8  & 44.0 \\
    Conv(dilation=2)   & 24.87M  & 79.7  & 74.5 & 63.1  & 42.9 \\
    Conv(dilation=3)   & 24.87M  & 79.9  & 74.1 & 62.9  & 42.6 \\
    \hline
    Deform Conv        & 24.91M  & 80.2  & 75.9 & 64.8  & 46.6 \\
    \hline
    Ours               &\textbf{24.87M}&\textbf{81.0}&\textbf{75.9}&\textbf{65.6}&\textbf{47.3}\\
    \hline
    \end{tabular}
    }
    \caption{Box regression capability of different attending processes over several IoU threshold.}
    \label{tab:box_localization}
\end{table}

\begin{table}[t]
\centering
    \resizebox{\linewidth}{!}{
    \begin{tabular}{c|c|c|ccc}
    \hline
    \ prediction module & params & m$AP$ & m$AP_{S}$ & m$AP_{M}$ & m$AP_{L}$ \\
    \hline
    \hline
    Conv               & 24.82M & 75.5 & 48.8 & 72.6 & 85.5 \\
    \hline
    Conv(1)-Conv       & 26.00M & 75.3 & 48.9 & 72.4 & 84.8 \\
    Conv(2)-Conv       & 27.19M & 75.0 & 48.7 & 71.6 & 85.0 \\
    \hline
    ResBlock(1)-Conv   & 27.19M & 75.2 & 50.6 & 71.1 & 84.5 \\
    ResBlock(2)-Conv   & 29.55M & 75.3 & 48.7 & 71.4 & 85.5 \\
    \hline
    PA                 & \textbf{24.87M}& \textbf{81.0} & \textbf{56.5} & \textbf{79.4} & \textbf{87.5} \\
    \hline
    \end{tabular}
    }
    \caption{Impact of increasing the depth of the standard convolutional prediction module. The number in parentheses denotes the number of intermediate layers of each type.}
    \label{tab:depth_ablation}
\end{table}


\section{Experiments}
We evaluate PASSD on two generic object detection benchmarks: PASCAL VOC 2007 and MS COCO. PASCAL VOC and MS COCO include 20 and 80 object classes, respectively. For benchmarks, we compare our model with other single-model entries under a single-scale evaluation for fair comparison.

\subsection{PASCAL VOC dataset}
\label{sec:pascal_voc}
We train our model on the union of the VOC 2007 \texttt{trainval} set and the VOC 2012 \texttt{trainval} set, and evaluate it on the VOC 2007 \texttt{test} set. The initial learning rate is 4$\times10^{-3}$ and divided by 10 at 150 and 200 epochs. The total number of training epochs is 250.

\subsubsection{Ablation Study}
In order to evaluate the effectiveness of our model, we conduct extensive ablation experiments. Moreover, to analyze the detection performance across the several sizes of objects, we evaluate the models using scale criteria of: \textit{small} (area $<64^2$), \textit{medium} ($64^2<$ area $<192^2$), and \textit{large} (area $>192^2$).

\medskip
\noindent\textbf{Proposing Process.}
To analyze the effectiveness of the proposing process, we add the process to the standard convolutional prediction module as shown in exp2 of~\tabref{tab:PA_ablation}. Adding the proposing process significantly improves the accuracy by a large margin of 3.5 points. In addition, clipping the box offset regression further pushes the accuracy as in exp3 of~\tabref{tab:PA_ablation}, demonstrating that diversifying the training samples is important. The results indicate that the proposing process is the key part for one-stage detectors.

\begin{figure}[t]
    \centering
    \begin{subfigure}[ht]{0.49\linewidth}
        \includegraphics[width=\textwidth]{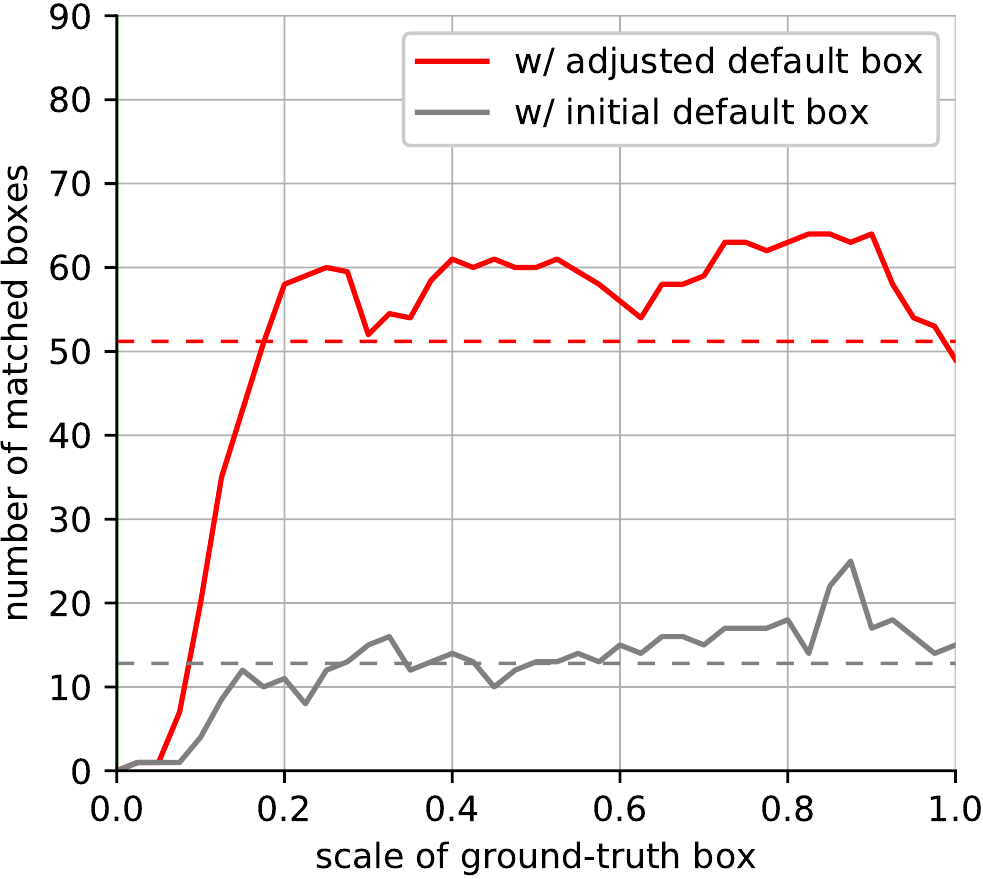}
        \label{fig:matched_anchor1}
    \end{subfigure}
    \hfill
    \begin{subfigure}[ht]{0.49\linewidth}
        \includegraphics[width=\textwidth]{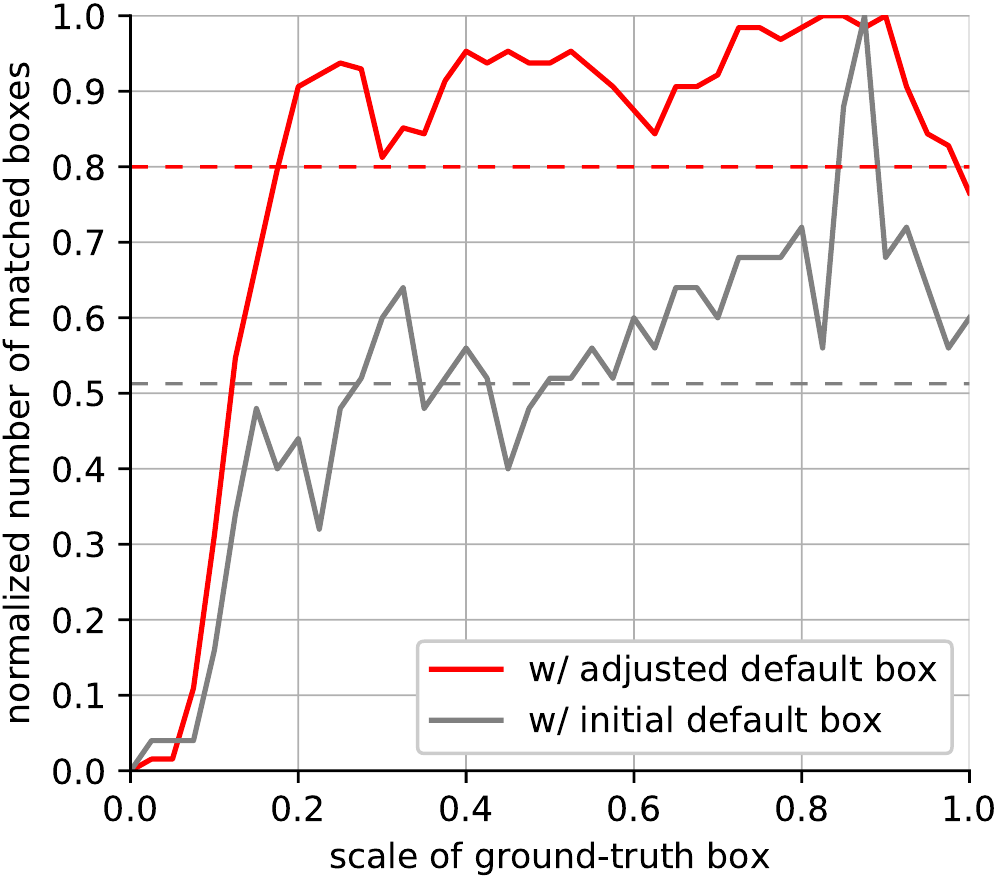}
        \label{fig:matched_anchor2}
    \end{subfigure}
    \vspace{-3mm}
    \caption{Impact of the proposing process. The left figure shows matched default boxes in terms of scale of ground-truth boxes. The right figure is a normalized version of the left figure. The average is represented by the dashed line.}
    \label{fig:matched_boxes}
\end{figure}

We also visualize the impact of the proposing process in~\figref{fig:matched_boxes}. We can clearly observe that the number of matched default boxes (\ie, positive training samples) significantly increased compared to the initial default box setting (\figref{fig:matched_boxes} (left)). Moreover, the normalized version (\figref{fig:matched_boxes} (right)) demonstrates that the proposing process distributes the positive training samples somewhat evenly across the object scales, effectively reducing the imbalance between them. To sum up, the proposing process provides a large number of well-balanced positive training samples during the training phase, which is especially effective for small and medium-sized objects.

\medskip
\noindent\textbf{Attending Process.}
For the attending process, we compare our approach with two different methods that can also adjust the sampling points of the convolution filter: dilated convolution~\cite{yu2017dilated} and deformable convolution~\cite{dai2017deformable}. Dilated convolution increases its sampling points with a \textit{fixed} discrete dilation value, whereas deformable convolution adjusts its sampling points in an \textit{adaptive} manner using an additional dedicated offset prediction layer, where the supervision signal for the offset prediction is given from the target task implicitly. The results are shown from exp4 to exp7 in~\tabref{tab:PA_ablation}. We observe that both dilated and deformable convolution have little impact on the accuracy in comparison to standard convolution (exp3~\tabref{tab:PA_ablation}). Whereas, our attending process brings a significant accuracy improvement over a wide range of object scales. Note that the accuracy gain does not come from the additional model capacity compared to standard convolution (exp3 in~\tabref{tab:PA_ablation}), demonstrating its effectiveness.

\medskip
\noindent\textbf{Propose-and-Attend Mechanism.}
Putting all together, our final model significantly improves the standard convolutional prediction module by a large margin of 5.5 points with marginal parameter overheads (0.05M), demonstrating that building the propose-and-attend mechanism itself, which is missed in most one-stage detectors, is indeed crucial for achieving a high detection accuracy. Note that the whole process is achieved in a single feed-forward manner without stage-discrimination as in two-stage detectors~\cite{ren2015faster}, ensuring high efficiency.

\medskip
\noindent\textbf{Box Regression Capability.}
To analyze the impact of the proposed attending process in final box regression, we compare our approach with other methods under different IoU thresholds. As shown in~\tabref{tab:box_localization}, our approach achieves better accuracy than all other methods over all IoU thresholds, demonstrating its higher box regression capability. Note that the accuracy gap between methods with \textit{fixed} sampling points and methods with \textit{adaptive} sampling points becomes larger as the IoU threshold increases, showing that \textit{dynamic} attend-mechanism, which is missed in most one-stage detectors, is essential for accurate box regression. Our approach successfully performs this mechanism without adding any parameter overheads to the standard convolution (first row in~\tabref{tab:box_localization}), revealing both the significance of the mechanism itself and the effectiveness of our approach.

\medskip
\noindent\textbf{Deeper Prediction Module.}
Here, we explore another direction in improving the standard convolutional prediction module,~\ie, increasing the depth of the prediction module. Specifically, we add two types of intermediate layers into the standard convolutional prediction module: 1) convolution and 2) residual block~\cite{he2016deep}, where the convolution intermediate layer is followed by ReLU activation. Both intermediate layers use 3$\times$3 filter and preserve the output channels to be same as the input feature (\ie, 256 in this work). Also, their parameters are independent for classification and box regression branch, respectively. As shown in~\tabref{tab:depth_ablation}, merely increasing the network depth has little improvement compared to the prediction module with a single convolution layer (first row of~\tabref{tab:depth_ablation}). Whereas, our PA prediction module shows significantly higher accuracy with marginal parameter overheads, demonstrating that building the propose-and-attend mechanism into the one-stage detector is indeed crucial.

\begin{table}[t]
    \centering
    \resizebox{\linewidth}{!}{
    \begin{tabular}{c|c|c|c|c|c|c}
        \hline
        model  &  backbone  &  input size  &  \# boxes  &  fps  &  params & m$AP$ \\
        \hline
        \hline
        \textit{Two-stage:}& & & & &\\
        Faster R-CNN~\cite{ren2015faster}        & VGG-16      & $\sim$ 1000$\times$600   & 300   & 7 & 135M & 73.2\\
        OHEM~\cite{shrivastava2016training}      & VGG-16      & $\sim$ 1000$\times$600   & 300   & 7 & -    & 74.6\\
        Faster R-CNN~\cite{ren2015faster}        & ResNet-101  & $\sim$ 1000$\times$600   & 300   & 2.4 & -  & 76.4\\
        R-FCN~\cite{dai2016rfcn}                 & ResNet-101  & $\sim$ 1000$\times$600   & 300   & 9  & 51M & 80.5\\
        Deep Regionlets~\cite{xu2018regionlets}       & ResNet-101  & $\sim$ 1000$\times$600   & 300   & -   & -& 82.0\\
        CoupleNet~\cite{Zhu_2017_CoupleNet}      & ResNet-101  & $\sim$ 1000$\times$600   & 300   & 8.2 & -& 82.7\\
        \hline
        \hline
        \textit{One-stage:}& & & & &\\
        SSD300~\cite{liu2016ssd}                 & VGG-16      & 300$\times$300      & 8732  & 46   & 27M   & 77.2\\
        YOLOv2~\cite{redmon2016yolo9000}         & Darknet-19  & 544$\times$544      & 845   & 40   & 67M   & 78.6\\
        DSSD321~\cite{fu2017dssd}                & ResNet-101  & 321$\times$321      & 17080 & 9.5  & -     & 78.6\\
        SSD512~\cite{liu2016ssd}                 & VGG-16      & 512$\times$512      & 24564 & 19   & 27M & 79.8\\
        RefineDet320~\cite{zhang2018refinedet}   & VGG-16      & 320$\times$320      & 6375  & 40.3 & -     & 80.0\\
        RFBNet300~\cite{liu2018rfb}              & VGG-16      & 300$\times$300      & 11620 & 83   & 37M   & 80.5\\
        SSD513~\cite{fu2017dssd}                 & ResNet-101  & 513$\times$513      & 43688 & 6.8  & -     & 80.6\\
        DSSD513~\cite{fu2017dssd}                & ResNet-101  & 513$\times$513      & 43688 & 5.5  & -     & 81.5\\
        RefineDet512~\cite{zhang2018refinedet}   & VGG-16      & 512$\times$512      & 16320 & 24.1 & -     & 81.8\\
        RFBNet512~\cite{liu2018rfb}              & VGG-16      & 512$\times$512      & 32756 & 38   & 37M   & 82.2\\
        \hline
        PASSD-320                            & VGG-16      & 320$\times$320  & 6375  & 50   & 25M & {81.0}\\
        PASSD-512                            & VGG-16      & 512$\times$512  & 16320 & 31.3 & 25M & {82.4}\\
        \hline
    \end{tabular}
    }
    \caption{Object detection results on PASCAL VOC 2007 \texttt{test} set.}
    \label{tab:voc2007}
\end{table}

\begin{table*}[!t]
\centering
\resizebox{\textwidth}{!}{%
\begin{tabular}{c|c|c|c|c|c|c|ccc|ccc}
    \hline
    model  &  data & backbone & input size & fps & params & size & $AP_{50}$ & $AP_{75}$ & $AP$ & $AP_{S}$ & $AP_{M}$  & $AP_{L}$  \\
    \hline
    \hline
    \textit{Two-stage:}& & & & & & & & & \\
    R-FCN~\cite{dai2016rfcn}                       & trainval   & ResNet-101               & $\sim$ $1000\times600$  & 9   & -   & 206MB & 51.9 & -    & 29.9   & 10.8 & 32.8 & 45.0 \\
    CoupleNet~\cite{Zhu_2017_CoupleNet}            & trainval   & ResNet-101               & $\sim$ $1000\times600$  & 8.2 & -   & -     & 54.8 & 37.2 & 34.4 & 13.4 & 38.1 & 50.8 \\
    Deformable R-FCN~\cite{dai2017deformable}      & trainval   & Aligned-Inception-ResNet & $\sim$ $1000\times600$  & -   & -   & -     & 58.0 & 40.8 & 37.5 & 19.4 & 40.1 & 52.5 \\
    Faster R-CNN w FPN~\cite{lin2017fpn}           & trainval35k& ResNet-101               & $\sim$ $1000\times600$  & 5.8 & 61M & 232MB & 59.1 & 39.0 & 36.2 & 18.2 & 39.0 & 48.2 \\
    Deep Regionlets~\cite{xu2018regionlets}        & trainval  & ResNet-101                & $\sim$ $1000\times600$  & -   & -   & -     & 59.8 & -    & 39.3 & 21.7 & 43.7 & 50.9 \\
    Mask R-CNN~\cite{he2017mask}                   & trainval35k\dag & ResNet-101  & $\sim$ $1280\times800$  & 4.8 & 63M  & 242MB & 60.3 & 41.7 & 38.2 & 20.1 & 41.1 & 50.2 \\
    Libra R-CNN~\cite{pang2019libra}               & trainval35k & ResNet-101              & $\sim$ $1280\times800$  & -   & 61M & 233MB & 62.1 & 44.7 & 41.1 & 23.4 & 43.7 & 52.5 \\
    Cascade R-CNN~\cite{cai2018cascade}            & trainval35k & ResNet-101              & $\sim$ $1280\times800$  & 7.1 & 88M & 337MB & 62.1 & 46.3 & 42.8 & 23.7 & 45.5 & 55.2 \\
    \hline
    \hline
    \textit{One-stage:}& & & & & & & & \\
    YOLOv2~\cite{redmon2016yolo9000}         & trainval35k & Darknet-19      & $416\times416$        & 40   & 67M  & -     & 44.0 & 19.2 & 21.6 & 5.0  & 22.4 & 35.5 \\
    SSD512~\cite{liu2016ssd}                 & trainval35k & VGG-16          & $512\times512$        & 22   & 36M  & 137MB & 48.5 & 30.3 & 28.8 & 10.9 & 31.8 & 43.5 \\
    RFBNet300~\cite{liu2018rfb}              & trainval35k & VGG-16          & $300\times300$        & -    & -    & -  & 49.3 & 31.8 & 30.3 & 11.8 & 31.9 & 45.9 \\
    RetinaNet500~\cite{lin2017focal}         & trainval35k & ResNet-101      & $\sim$ $832\times500$ & 11.1 & 57M  & 217MB & 53.1 & 36.8 & 34.4 & 14.7 & 38.5 & 49.1 \\ 
    RFBNet512~\cite{liu2018rfb}              & trainval35k & VGG-16          & $512\times512$        & -    & 47M  & -     & 54.2 & 35.9 & 33.8 & 16.2 & 37.1 & 47.4 \\
    RefineDet512~\cite{zhang2018refinedet}   & trainval35k & VGG-16          & $512\times512$        & 22.3 & -    & 137MB & 54.5 & 35.5 & 33.0 & 16.3 & 36.3 & 44.3 \\
    ExtremeNet~\cite{zhou2019extnet} (flip)  & trainval35k & Hourglass-104   & $511\times511$        & 3.1  & -    & 758MB & 55.5 & 43.2 & 40.2 & 20.4 & 43.2 & 53.1 \\
    RFBNet512-E~\cite{liu2018rfb}            & trainval35k & VGG-16          & $512\times512$        & -    & 59M  & 191MB & 55.7 & 36.4 & 34.4 & 17.6 & 37.0 & 47.6 \\
    CornerNet~\cite{law2018corner} (flip)    & trainval35k & Hourglass-104   & $511\times511$        & 4.1  & 201M & 768MB & 56.5 & 43.1 & 40.5 & 19.4 & 42.7 & 53.9 \\
    RefineDet512~\cite{zhang2018refinedet}   & trainval35k & ResNet-101      & $512\times512$        & -    & -    & 315MB & 57.5 & 39.5 & 36.4 & 16.6 & 39.9 & 51.4 \\
    RetinaNet800~\cite{lin2017focal}         & trainval35k & ResNet-101  & $\sim$ $1280\times800$    & 5.1  & 57M  & 217MB & 59.1 & 42.3 & 39.1 & 21.8 & 42.7 & 50.2 \\
    M2Det~\cite{Zhao2019m2det}               & trainval35k & VGG-16          & $800\times800$        & 11.8 & 147M & 506MB & 59.7 & 45.0 & 41.0 & 22.1 & 46.5 & 53.8 \\
    \hline
    PASSD-320                           & trainval35k & VGG-16          & $320\times320$  & 40   & 25M & 96MB & 51.6 & 33.6 & 31.4 & 12.0 & 35.1 & 45.8 \\
    PASSD-512                           & trainval35k & VGG-16          & $512\times512$  & 22.2 & 25M & 96MB & 56.9 & 38.4 & 35.3 & 19.2 & 39.0 & 45.5 \\
    PASSD-320                           & trainval35k & ResNet-101      & $320\times320$  & 34.5 & 47M & 181MB & 52.1 & 35.3 & 32.7 & 10.8 & 36.5 & 50.2 \\
    PASSD-512                           & trainval35k & ResNet-101      & $512\times512$  & 22.2 & 47M & 181MB & 59.1 & 41.4 & 37.8 & 19.3 & 42.6 & 51.0 \\
    PASSD-768                           & trainval35k & ResNet-101      & $768\times768$  & 11.9 & 48M & 184MB & 62.1 & 44.7 & 40.3 & 24.2 & 44.8 & 50.3 \\
    \hline
\end{tabular}%
}
\caption{Object detection results on MS COCO \texttt{test-dev} set. "\dag" denotes the use of additional pixel-level supervision. "flip" indicates that the model is evaluated on both original and flipped input image.}
\label{tab:coco}
\end{table*}

\vspace{-5pt}
\subsubsection{Comparison to State of the Art}
We compare our final model with the state-of-the-art detection models in~\tabref{tab:voc2007}. PASSD with low resolution input (\ie, 320$\times$320) achieves 81.0 mAP. This result is much better than those of several two-stage methods, such as R-FCN~\cite{dai2016rfcn}, which uses a larger input size and a deeper backbone (\ie, ResNet-101). With a larger input (\ie, 512$\times$512), PASSD produces 82.4 mAP, surpassing all detection models, including both one-stage and two-stage, except CoupleNet~\cite{Zhu_2017_CoupleNet} with a marginal gap (0.3 mAP). Note that CoupleNet uses a larger input size ($\sim$1000$\times$600) and adopts a deeper backbone (~\ie, ResNet-101) than PASSD-512. Compared to other one-stage detectors, such as SSD or DSSD, our model achieves better accuracy with fewer default boxes (\eg, 43,688 default boxes in DSSD513 vs. 16,320 default boxes in PASSD-512). This implies that PASSD can handle various object scales effectively. Finally, our model uses fewer parameters than almost any other models, showing that the superiority of our model does not come from the mere high model capacity, but from the effective architecture design.

We also report the inference time of our model in the fifth column of~\tabref{tab:voc2007}. The inference time is evaluated with a batch size of 1 on NVIDIA Titan X GPU, CUDA 8.0, and cuDNN v7. PASSD can process an image in 20 ms (50 fps) and 32 ms (31.3 fps) with input sizes of 320$\times$320 and 512$\times$512, respectively. While it is hard to perform apple-to-apple comparisons due to inconsistent environments (~\ie, different hardware and software libraries), PASSD shows real-time capability.

    

\begin{table}[t]
\centering
\resizebox{\linewidth}{!}{%
\begin{tabular}{c|c|c|c|ccc}
    \hline
    model                                    & backbone     & fps & params  & $AP$    & $AP_{50}$ & $AP_{75}$ \\
    \hline
    \hline
    YOLOv2-416~\cite{redmon2016yolo9000}         & DarkNet-19   & 40 & 67.4M    & 21.6  & 44.0    & 19.2    \\
    SSD-300~\cite{liu2016ssd}                    & VGG-16       & 43 & 34.3M    & 25.1  & 43.1    & 25.8    \\
    \hline
    SSD-300~\cite{liu2016ssd}                    & MobileNet    & 80    & 6.8M   & 18.8  & -     & -  \\
    SSDlite-300~\cite{Sandler2018mobilenetv2}    & MobileNet v2 & 61    & 4.3M  & 22 & - & - \\
    \hline
    PASSD-320                                 & MobileNet    & 63 & 6.7M & 25.3 & 43.6 & 26.3    \\
    \hline
    
\end{tabular}%
}
\caption{Object detection results with lightweight backbone on MS COCO \texttt{test-dev}.}
\label{tab:lightweight_backbone}
\end{table}

\begin{table}[t]
\centering
\resizebox{0.95\linewidth}{!}{%
\begin{tabular}{c|c|c|c}
    \hline
    model  &   backbone      & prediction module  &  m$AP$   \\
    \hline
    \hline
    DSOD-300~\cite{Shen_2017_dsod}          & DS/64-192-48-1 & Conv  & 77.7  \\
    ScratchDet-300~\cite{Zhu2019scratchdet} & Root-ResNet-18 & Conv  & 78.5  \\
    \hline
    PASSD-320                              & VGG-16-BN      & Conv  & 73.2  \\
    PASSD-320                              & VGG-16-BN      & PA    & 79.1  \\
    \hline
\end{tabular}%
}
\caption{Comparison to other train-from-scratch models on VOC 2007 \texttt{test} set.}
\label{tab:scratch}
\end{table}
\vspace{-3pt}

\subsection{MS COCO dataset}
\label{sec:ms_coco}
To further validate the proposed PASSD in a large-scale setting, we evaluate our model on MS COCO. We also report the results of the model using the ResNet-101 backbone to see the effect of adopting a deeper backbone. We train our model on~\texttt{trainval35k} and report the main results on~\texttt{test-dev}. The initial learning rate is 2$\times10^{-3}$ and divided by 10 at 80 and 100 epochs. The total number of training epochs is 120.

\subsubsection{Comparison to State of the Art}
The results are shown in~\tabref{tab:coco}. PASSD achieves 31.4 AP with an input size of 320$\times$320 and VGG-16 backbone. The accuracy of PASSD is further improved by 35.3 AP when a larger input size (\ie, 512$\times$512) is used. Meanwhile, adopting a deeper backbone (\ie, ResNet-101) further pushes the accuracy of PASSD; it results in 32.7 AP, 37.8 AP, and 40.3 AP for 320$\times$320, 512$\times$512, and 768$\times$768 input sizes respectively. The PASSD-768 achieves results competitive to state-of-the-art models by adding only marginal parameter overheads to the backbone network, resulting in a much lighter model than competitive approaches. This shows that the superiority of our model does not come from the mere high model capacity, but from the effective architecture design. In particular, our best model shows state-of-the-art accuracy on $AP_{50}$ and $AP_{S}$, and it occupies top-entries on $AP_{75}$. It also runs faster than most competitive methods. The results indeed demonstrate the effectiveness of the proposed method. In addition, recent ideas of designing a better backbone (M2Det~\cite{Zhao2019m2det}), applying a better training procedure (Libra R-CNN~\cite{pang2019libra}), and applying the prediction module in a cascade manner (Cascade R-CNN~\cite{cai2018cascade}) are orthogonal to our approach of designing an effective prediction module, having potential to be used in a complementary manner.

\medskip
\subsection{Discussion}
\subsubsection{Lightweight Backbone}
Our final model features a simple design with marginal parameter overheads to the backbone network. Therefore, we apply our propose-and-attend (PA) prediction module to a lightweight backbone to further improve efficiency. We train our PASSD with MobileNet~\cite{howard2017mobilenets} as the backbone on MS COCO with the same training setting. As shown in~\tabref{tab:lightweight_backbone}, PASSD significantly outperforms other lightweight detectors, even surpassing the models with advanced backbone such as DarkNet-19~\cite{redmon2016yolo9000} and VGG-16~\cite{simonyan2014very}. Moreover, recent works on designing a better lightweight backbone (\eg, MobileNet v2~\cite{Sandler2018mobilenetv2}) are complementary to our approach of designing the better prediction module. The result demonstrates its great potential for low-end devices.

\subsubsection{Training from Scratch}
We also observe that a model integrated with the proposed PA prediction module can be trained from scratch (\ie w/o ImageNet pretraining). Recently, Shen et al.~\cite{Shen_2017_dsod} showed that training one-stage detectors without a pretrained backbone network is hard. To address this, recent approaches have attempted to carefully design a backbone network that is suitable for this setting. Apart from recent approaches~\cite{Shen_2017_dsod,Zhu2019scratchdet}, we found that simply integrating our proposed PA prediction module into the network enables successful training. As seen in~\tabref{tab:scratch}, we achieved favorable results by only inserting batch normalization~\cite{Ioffe2015BN} in the backbone network without using any sophisticated backbone design. Note that without the PA prediction module, the performance significantly drops, implying that the PA prediction module indeed provides a rich supervisory signal during training.

\section{Conclusion}
We present PASSD, a novel one-stage detector that functions with the \textit{propose-and-attend} mechanism efficiently. We conduct extensive ablation studies to validate the efficacy of the proposed method. We evaluate our model on several benchmarks and show results competitive to state-of-the-art models while using much fewer parameters thanks to our efficient design. Moreover, we demonstrate that our method can be successfully applied to the lightweight backbone network and train-from-scratch scheme. We believe the experimental results and analysis provided in this paper would benefit the community and the practitioners.


\clearpage

{\small
\bibliographystyle{ieee}
\bibliography{egbib}
}

\end{document}